\documentclass{article}

\usepackage{amsmath, amssymb, amsthm, algorithm, algpseudocode, subcaption}
\usepackage{graphicx}

\newcommand{\delt}[1]{{\scriptsize\textcolor{gray}{(#1)}}}
\newcommand{\std}[1]{{\scriptsize\textcolor{gray}{$\pm$#1}}}

\newenvironment{denseitemize}{
\begin{itemize}[topsep=2.5pt, partopsep=0pt, leftmargin=1.5em]
  \setlength{\itemsep}{2.5pt}
  \setlength{\parskip}{0pt}
  \setlength{\parsep}{0pt}
}{\end{itemize}}

\usepackage{pifont}
\newcommand{\cmark}{\textcolor{green!60!black}{\ding{51}}}
\newcommand{\xmark}{\textcolor{red!70!black}{\ding{55}}}

 \usepackage[main, final]{neurips_2026}

\usepackage[utf8]{inputenc} %
\usepackage[T1]{fontenc}    %
\usepackage{hyperref}       %
\hypersetup{
  colorlinks,
  linkcolor={red!70!black},
  citecolor={blue!70!black},
  urlcolor={blue!70!black}
}
\usepackage{cleveref}
\usepackage{url}            %
\usepackage{booktabs}       %
\usepackage{amsfonts}       %
\usepackage{nicefrac}       %
\usepackage{microtype}      %
\usepackage{xcolor}         %
\usepackage{enumitem}

\title{TriAxialKV: Toward Extreme Low-Precision KV-Cache Quantization for Agentic Inference Tasks}

\author{%
  Hanzhang Shen$^1$ $\quad$
  Haoran Wu$^1$ $\quad$
  Yiren Zhao$^2$ $\quad$
  Robert Mullins$^{1,}$\thanks{Corresponding author: \href{mailto:Robert.Mullins@cl.cam.ac.uk}{\color{blue!70!black}Robert.Mullins@cl.cam.ac.uk}} \\
  $^1$ University of Cambridge \quad $^2$ Imperial College London
}

\begin{document}

\maketitle

\begin{abstract}

Agentic workloads have emerged as a major workload for LLM inference. They differ significantly from chat-only workloads, requiring long-context processing, the ability to handle multimodal inputs, and structured multi-turn interactions with tool calling capabilities.
As a result, their context exhibits structure that can carry different importance along three key axes: temporal recency to the current turn, modality such as text or image tokens, and semantic role such as user queries, tool calls, observations, or reasoning. These axes capture distinct token behaviors and lead to different sensitivities to KV-cache compression.
However, existing KV-cache quantization methods are typically homogeneous or exploit only heterogeneity on a single dimension, such as temporal proximity or modality, overlooking the interactions among them.
To this end, we introduce \textbf{TriAxialKV}, a novel mixed-precision KV-cache quantization scheme that assigns each token a triaxial tag, calibrates per-tag sensitivity, and allocates INT2/INT4 bit-widths under a fixed memory budget.
We implement TriAxialKV as an end-to-end serving system, comprising calibration, mixed-precision quantization and memory management, and custom fused Triton decode kernels.
When using Qwen3-VL-32B-Thinking as a computer-use agent operating the OSWorld, TriAxialKV matches the accuracy of SGLang with BF16 KV cache while supporting 4.5$\times$ KV cache capacity and achieving 30\% higher end-to-end throughput, when running on real GPU systems.

\end{abstract}

\section{Introduction}
Large language models (LLMs) have rapidly evolved from conversational assistants~\citep{wang2025burstgpt,lmsys-chat,banerjee2023benchmarkingllmpoweredchatbots} to agentic applications such as web search~\citep{webagent,he2024webvoyager}, computer use~\citep{agent_s2,owworld_human}, and code generation~\citep{humaneval, kernelcraft,rando2025longcodebench}, where the model autonomously plans, invokes tools, consumes multimodal inputs, and reasons over feedback across many turns.
These workloads require the model to attend over very long, multi-turn contexts, causing the key--value (KV) cache to grow rapidly and become a dominant memory bottleneck in LLM inference. As the KV cache expands, it can quickly saturate GPU memory, thereby limiting the achievable batch size and throughput~\citep{memory_walls,davies2025liminalexploringfrontiersllm}.
For instance, a 100K-token context with Qwen3-VL-32B~\citep{Qwen3-VL} requires approximately 24~GiB of KV-cache memory at BF16 precision, even for a single request.
Combined with model weights and runtime buffers, such large KV caches leave substantially less GPU memory for concurrent requests, limiting batch size and serving throughput.

KV-cache compression is a promising approach to improve agentic serving efficiency. A wide range of methods have been proposed, including uniform low-bit quantization~\citep{sawint4,quarot}, asymmetric quantization~\citep{liu2024kivi,tao2025asymkv}, token-eviction strategies that exploit attention sparsity~\citep{zhang2023h2o,zhaosmallkv}, and adaptive bit-width allocation along the layer depth or generation axis~\citep{boroujeni2026dontwastebitsadaptive,liu2026pmkvq}.
However, these methods primarily adapt compression according to quantization statistics, attention-based importance, or individual structural properties of the KV cache, without fully accounting for the heterogeneous nature of agentic workloads.
In particular, agentic contexts exhibit substantial heterogeneity along three dimensions: tokens can (i) originate from different \emph{modalities} (e.g., text, image), (ii) arrive at different \emph{times} in the conversation (system prompts, early user turns, recent tool outputs), and (iii) play different \emph{semantic} roles (instructions, reasoning, tool-call arguments, tool responses).

\begin{figure}[t]
    \centering
    \includegraphics[width=0.85\linewidth]{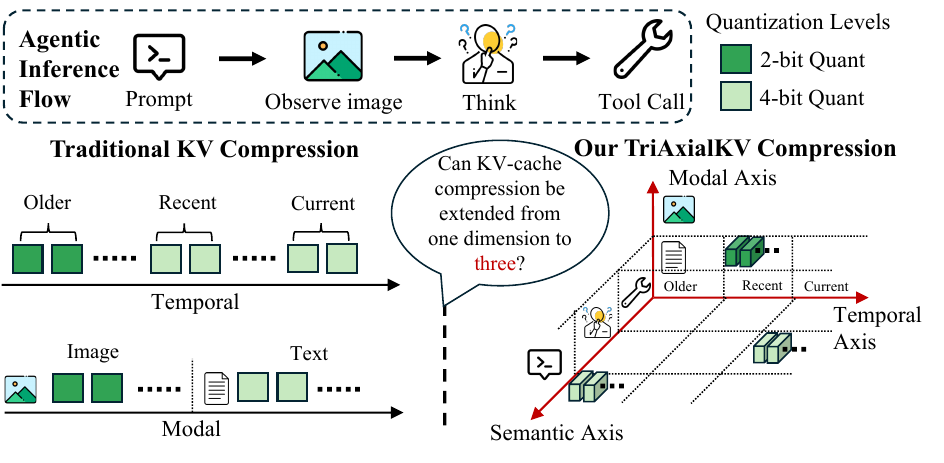}
    \caption{Comparison of single-axis KV-cache compression methods, including PM-KVQ~\citep{liu2026pmkvq} for temporal compression and VL-Cache~\citep{tu2025vlcache} for modality-aware compression, with TriAxialKV, which holistically compresses the KV cache by jointly considering temporal, modal, and semantic axes.}
    \label{fig:motivation}
\end{figure}

Recent works have begun to exploit individual dimensions for KV-cache compression.
ThinKV~\citep{thinkv} considers the semantic role of tokens, VL-Cache~\citep{tu2025vlcache} and LOOK-M~\citep{wan2024lookm} classify tokens based on modality, and PM-KVQ~\citep{liu2026pmkvq} categorizes reasoning tokens according to temporal recency.
However, as illustrated in \Cref{fig:motivation}, they \emph{treat these dimensions in isolation rather than jointly modeling the temporal, modal, and semantic structure of agentic traces}.
Consequently, existing compression policies cannot distinguish tokens that share the same tag along one axis but differ in quantization sensitivity along the others, leading to accuracy degradation under aggressive compression.

To address this gap, we propose \textbf{TriAxialKV}, a mixed-precision KV-cache quantization framework that assigns each token a triaxial tag based on its temporal recency, modality, and semantic role.
By calibrating the quantization sensitivity of each tag, TriAxialKV allocates INT2/INT4 precision under a memory budget, enabling aggressive KV-cache compression while preserving model accuracy.
In summary, our main contributions are:
\begin{denseitemize}
    \item By profiling function-calling and computer-use prefills, we show that KV quantization sensitivity is structured along three complementary axes: \emph{temporal recency}, \emph{modality}, and \emph{semantic role}. To our knowledge, this is the \emph{first} work to exploit all three jointly for KV-cache quantization.
    \item Building on this finding, we develop a \emph{taxonomy-driven calibration and allocation framework}: a chat-template-based tagger that labels every prefill token in a single pass without any model inference, a calibration procedure that measures per-tag attention-output distortion from real prefill captures, and a per-tag bit-width allocator that solves the INT2/INT4 assignment problem.
    \item We deliver an \emph{end-to-end serving system} that integrates the tagger, allocator, and mixed-precision memory pool with fused Triton decode kernels into SGLang~\citep{zheng2024sglang}. On Qwen3-VL-32B~\citep{Qwen3-VL} running OSWorld~\citep{OSWorld}, our system provides \emph{4.5$\times$ the KV-cache capacity} and serves \emph{1.3$\times$ the end-to-end throughput} of the BF16 baseline on an NVIDIA B200 GPU while matching its accuracy. On BFCL Memory~\citep{patil2025bfcl}, it stays within 2\% of BF16 accuracy across four models, while KIVI~\citep{liu2024kivi} and SGLang FP4 lose up to 6.0\% and 4.7\%.
\end{denseitemize}

\section{Related Work}
\paragraph{KV cache compression}
To reduce the memory footprint of long-context inference and enable larger batch sizes, prior work has extensively explored KV cache quantization.
KIVI~\citep{liu2024kivi} employs two-bit per-channel quantization for keys and per-token quantization for values to account for their distributions.
KVQuant~\citep{hooper2024kvquant} enables sub-4-bit precision through pre-RoPE key quantization, non-uniform datatypes, and outlier isolation.
QServe~\cite{lin2025qserve} co-designs W4A8KV4 quantization with an efficient serving system to translate memory savings into throughput improvements.
OSCAR~\citep{zhou2026oscarofflinespectralcovarianceaware} calibrates per-layer rotations and clipping thresholds offline from attention-aware KV covariance, enabling 2-bit KV quantization with a small BF16 sink and recent window.

Other approaches exploit variations in KV cache sensitivity to allocate precision non-uniformly.
ZipCache~\citep{he2024zipcache} identifies salient tokens via channel-separable scoring to enable aggressive compression of less-important tokens.
PM-KVQ~\citep{liu2026pmkvq} employs progressive mixed-precision
quantization to reduce the bit-width of KV cache blocks.
ThinKV~\citep{thinkv} adapts quantization and eviction to the importance of different reasoning thoughts, whereas ChanMix~\citep{chanmix} allocates precision according to channel sensitivity.
While these methods demonstrate the benefits of sensitivity-aware quantization, our work jointly exploits temporal recency, modality, and semantic role to guide precision allocation for agentic workloads.
A complementary line of work explores token eviction for KV-cache compression~\citep{zhang2023h2o,xiao2023streamingllm, ge2023fastgen,li2024snapkv,cai2024pyramidkv,fengada,zhaosmallkv}.
Token eviction is complementary to our approach, as it reduces the number of retained tokens, whereas our method reduces the precision of retained KV entries.

\paragraph{Efficient LLM serving systems}
A substantial body of work has addressed system-level challenges in LLM serving.
Orca~\citep{280922} introduces iteration-level scheduling
and selective batching to improve GPU utilization.
Building on iteration-level batching, Sarathi-Serve~\citep{agrawal2024taming} introduces chunked prefills and stall-free scheduling to balance throughput and tail latency.
FlashAttention~\citep{dao2022flashattention, dao2023flashattention2, shah2024flashattention, zadouri2026flashattention4algorithmkernelpipelining} reduces memory traffic and optimizes GPU kernel execution for efficient attention computation.
vLLM~\citep{kwon2023vllm} uses PagedAttention to manage KV cache in non-contiguous blocks and reduce memory fragmentation.
SGLang~\citep{zheng2024sglang} employs a radix tree to organize KV cache and enable prefix reuse among requests.
Another line of work explores prefill--decode
disaggregation, as exemplified by DistServe~\citep{zhong2024distserve} and Splitwise~\citep{patel2024splitwise}.
Mooncake~\citep{qin2025mooncake} further incorporates disaggregated KV cache storage and cache-aware scheduling.
Our work complements these system-level optimizations by reducing KV cache memory consumption through sensitivity-aware quantization.
We integrate our approach into SGLang to demonstrate end-to-end serving throughput improvements.

\section{Method}
\label{method}

\begin{figure}[t]
  \centering
  \begin{subfigure}[b]{0.52\linewidth}
    \centering
    \includegraphics[width=\linewidth]{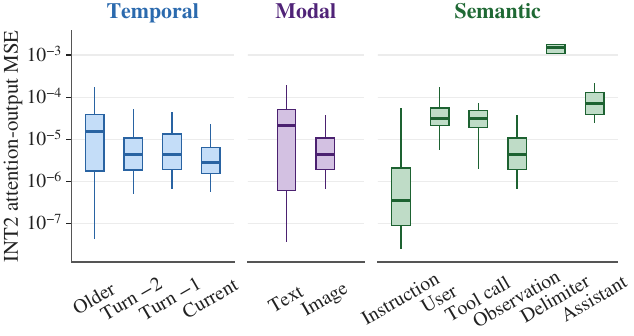}
    \caption{Qwen3-VL-8B, OSWorld}
    \label{fig:axis_spread_osworld}
  \end{subfigure}%
  \hfill
  \begin{subfigure}[b]{0.46\linewidth}
    \centering
    \includegraphics[width=\linewidth]{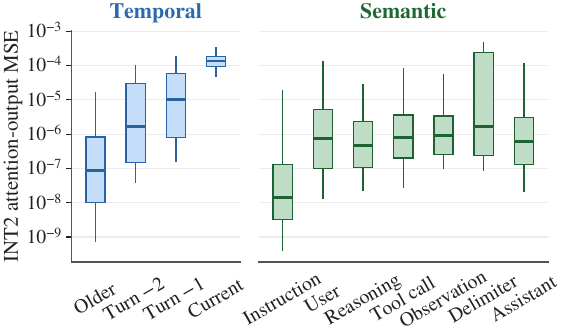}
    \caption{Qwen3-32B, BFCL}
    \label{fig:axis_spread_bfcl}
  \end{subfigure}
  \vspace{-1mm}
  \caption{INT2 quantization sensitivity of 32-token same-tag groups along each tag axis. Each group's KV is quantized to INT2 alone, with all other KV entries kept in BF16.}
  \label{fig:axis_spread}
\end{figure}

\begin{figure}[t]
    \centering
    \includegraphics[width=1\linewidth]{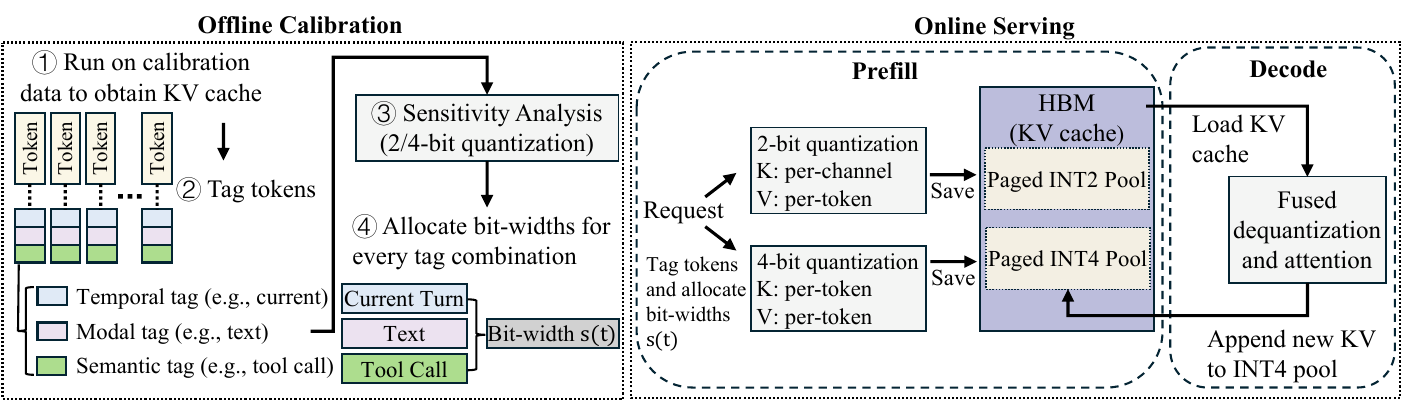}
    \caption{Overview of our TriAxialKV compression flow. During prefill, KV entries are quantized and stored in INT2/INT4 pools based on the bit-width assigned to the token's triaxial tag.}
    \label{fig:system}
\end{figure}

An agentic prefill request consists of heterogeneous segments, such as the system prompt, user queries, tool calls, and tool outputs.
To measure their quantization sensitivity, we tag every prefill token by its \textit{temporal recency}, \textit{modality}, and \textit{semantic role}, and split the tokens sharing all three tags into groups of 32.
We quantize one group at a time to INT2, keeping all else in BF16, and record the attention-output mean-squared error (MSE) at a middle layer.
As shown in \Cref{fig:axis_spread}, semantic role separates segments on both workloads, with instruction tokens one to two orders of magnitude less sensitive than most other roles.
Temporal recency matters more on BFCL, where sensitivity rises monotonically by over two orders of magnitude toward the current turn.
Text and image tokens differ by only about $5\times$, but images make up 84\% of OSWorld (computer-use) prefill tokens, so assigning them INT2 frees budget to keep more sensitive tokens at INT4.

This structure enables a \emph{sensitivity-aware allocation} that assigns a bit-width to each tag combination, which we call a \emph{triaxial tag}, subject to an average bit-width budget.
As shown in \Cref{fig:system}, our system has two stages.
First, an offline calibration stage measures the attention-output distortion of quantizing each triaxial tag to INT2 and INT4, and solves for a bit-width allocation under the budget.
Second, at serving time, the system tags each prefill token and looks up the bit-width of its triaxial tag.
We maintain separate paged INT2 and INT4 memory pools, and implement a fused Triton decode kernel that runs Flash-Decoding~\citep{dao2023flashdecoding} with on-the-fly dequantization of both bit-widths.

\subsection{Triaxial Tagging and Bit-Width Allocation}
\label{sec:allocation}

\paragraph{Tag taxonomy.}
\label{sec:taxonomy}
Our \emph{tag taxonomy} gives each prefill token $t$ one value on each of three axes: temporal, modal, and semantic.
The tuple of these values is its triaxial tag $s(t)$ in the joint tag space
\begin{equation}
\mathcal{S} = \mathcal{A}_{\mathrm{temporal}} \times \mathcal{A}_{\mathrm{modal}} \times \mathcal{A}_{\mathrm{semantic}},
\end{equation}
where each axis captures a distinct property of the segment.

\paragraph{Temporal.} $\mathcal{A}_{\mathrm{temporal}} = \{\mathtt{older}, \mathtt{turn\_m2}, \mathtt{turn\_m1}, \mathtt{current}\}$.
We group tokens by their distance from the most recent turn, where a turn spans from one user message up to, but not including, the next.
$\mathtt{current}$ covers the most recent turn, $\mathtt{turn\_m1}$ and $\mathtt{turn\_m2}$ cover the two turns before it, and $\mathtt{older}$ covers everything earlier.
Because attention to a turn tends to change as the conversation moves on, tokens from different turns can differ in quantization sensitivity.

\paragraph{Modal.} $\mathcal{A}_{\mathrm{modal}} = \{\mathtt{text}, \mathtt{image}\}$.
Image tokens enter the language model from the vision encoder rather than the embedding table, so their KV can have different statistics, such as per-channel variance and sparsity, and can respond differently to low-bit quantization.

\paragraph{Semantic.} $\mathcal{A}_{\mathrm{semantic}} = \{\mathtt{inst}, \mathtt{user}, \mathtt{assistant}, \mathtt{reasoning}, \mathtt{tool\_call}, \mathtt{obs}, \mathtt{delim}\}$.
These are functional roles in the agent loop, summarized in Table~\ref{tab:semantic-axis}.
\begin{table}[t]
\vspace{-4mm}
\centering
\small
\caption{Semantic axis values in our tag taxonomy.}
\begin{tabular}{ll}
\toprule
Semantic Tag & Description \\
\midrule
$\mathtt{inst}$      & system prompt and tool schemas \\
$\mathtt{user}$      & user-message text (excluding tool outputs) \\
$\mathtt{assistant}$ & assistant text outside reasoning and tool-call blocks \\
$\mathtt{reasoning}$ & chain of thought inside \texttt{<think>} blocks \\
$\mathtt{tool\_call}$ & tool invocations inside \texttt{<tool\_call>} blocks \\
$\mathtt{obs}$       & environment feedback (tool outputs, screenshots, etc.) \\
$\mathtt{delim}$     & role markers and the open/close tokens of the regions above \\
\bottomrule
\end{tabular}

\label{tab:semantic-axis}
\end{table}
Tags that never occur, such as $(\cdot, \mathtt{image}, \mathtt{reasoning})$, have zero token count and drop out of the optimization.
All three axes are determined by the chat template alone, without inspecting token content or running the model, so the tagger runs cheaply during request preprocessing.

\subsubsection{Allocation Setup}
Consider a request whose prefill contains $N$ tokens.
Each token $t$ contributes a key vector and a value vector to the KV cache, and receives a tag $s(t) \in \mathcal{S}$.
We assign each tag $k$ a bit-width $b_k \in \{2, 4\}$ shared by all tokens with that tag.
Our goal is to minimize the perturbation of the attention output subject to a target average bit-width over the prefill tokens.

\subsubsection{Allocation Objective}
Let $o_i$ denote the full-precision attention output at query position $i$, and $\tilde{o}_i(\mathbf{b})$ the output when KV vectors are quantized according to a per-tag bit-width allocation $\mathbf{b} = (b_k)_{k \in \mathcal{S}}$.
We aim to find the allocation $\mathbf{b}$ that minimizes the expected MSE of attention output:
\begin{equation}
  \mathcal{L}(\mathbf{b}) \;\triangleq\; \mathbb{E}_i\!\left[\,\|o_i - \tilde{o}_i(\mathbf{b})\|_2^2\,\right].
  \label{eq:objective}
\end{equation}
We target the attention output MSE rather than KV reconstruction MSE because the two are not monotonically related: a token with small KV quantization error can still contribute large output error if attention concentrates on it, whereas a large error on a rarely attended token has little effect.
To make this MSE tractable, we expand the attention output to first order in the quantization noise $\delta$, yielding an expression linear in the per-token noise.
We assume $\delta$ is zero-mean and uncorrelated across tags.
The cross-tag terms then vanish in expectation, and the loss decomposes over tags:
\begin{equation}
\small
  \hat{\mathcal{L}}(\mathbf{b}) = \sum_{k \in \mathcal{S}} D_k(b_k),
  \label{eq:decomp}
\end{equation}
where $D_k(b)$ is the output MSE when only tokens with tag $k$ are quantized to $b$ bits and all other tokens remain full precision.
Under this approximation, $D_k(b)$ depends only on the tag $k$ and its assigned bit-width $b$.
We measure $D_k(b)$ directly from calibration data: on a calibration capture of KV cache, we quantize only tokens with tag $k$ at bit-width $b$, replay attention, and record the output MSE against the full-precision reference.
Therefore, our analysis produces a table of $2 \cdot |\mathcal{S}|$ distortion values, one per tag and bit-width, which is reused across bit-width budgets during optimization.

\subsubsection{Optimization}
\label{sec:optim}
Let $N_k = |\{ t:s(t)=k \}|$ denote the token count for tag $k$ in the KV cache.
We estimate $N_k$ in two ways.
For $\mathtt{inst}$ tags, which cover the system prompt and tool schemas, tokens are shared across most, if not all, requests under prefix caching, and the cache stores a single copy.
Therefore, we take $N_k$ as the median token count in $\mathtt{inst}$-semantic segments across calibration requests.
For all other tags, we sum the prefill token count across all calibration requests.
With target average bit-width $B \in [2, 4]$, the allocation problem is:
\begin{equation}
\small
  \min_{\mathbf{b} \in \{2,4\}^{|\mathcal{S}|}}\;\; \sum_{k \in \mathcal{S}} D_k(b_k)
  \qquad \text{s.t.} \qquad \sum_{k \in \mathcal{S}} N_k\, b_k \;\le\; B \sum_{k \in \mathcal{S}} N_k.
  \label{eq:allocation}
\end{equation}
Upgrading tag $k$ from 2 to 4 bits costs $2N_k$ extra bits in the cache and reduces the total MSE by $D_k(2)-D_k(4)$.
We define the per-bit gain of this upgrade as:
\begin{equation}
\small
  \rho_k \;\triangleq\; \frac{D_k(2) - D_k(4)}{2\, N_k}.
  \label{eq:rho}
\end{equation}
When at most 22 tags are active ($N_k \geq 1$), we enumerate all allocation configurations and select the one with the minimal total MSE.
For larger numbers of active tags, we take a greedy approach: sort tags by $\rho_k$ in descending order, walk the sorted list, and upgrade each tag whose extra cost $2N_k$ fits in the remaining budget.
Algorithm~\ref{alg:alloc} in Appendix~\ref{sec:greedy_allocation} describes our greedy allocation.

\subsubsection{Calibration}
\label{sec:calibration}
For each model and target workload, we take 5\% of the dataset to form a calibration set, and use it to compute the per-tag bit-width allocation in three stages.

\paragraph{KV capture.}
We run the calibration set through the serving system to capture the KV traces.
For each model, we select four to six layers, evenly spread across the model's depth, and register forward hooks on each layer's attention module, which receives $Q$, $K$, and $V$ after RoPE~\citep{roformer}.
The hooks fire on prefill only, capturing $Q$ for the new tokens and $KV$ of the prefill sequence.

\paragraph{Sensitivity analysis.}
For each capture and each tag $k$ active in that capture, we quantize the KV of tokens tagged with $k$ using bit-width $b\in\{2, 4\}$, compute the attention output, and record the MSE against the full-precision reference.
This yields a raw $D_k^{\ell, r, h}(b)$ for every layer $\ell$, request $r$, and head $h$.
We aggregate these into a final sensitivity score $D_k(b)$ as follows:
\begin{enumerate}[leftmargin=1.5em]
    \vspace{-1mm}
    \item \textbf{Max across heads.} We take the maximum error across heads to ensure that no single attention subspace becomes a bottleneck, keeping the most sensitive head functional.
    \vspace{-1mm}
    \item \textbf{Mean across requests.} We average the results over all calibration requests to ensure the metric is representative and stable across different inputs.
    \vspace{-1mm}
    \item \textbf{Sum across layers.} We sum the errors across all layers to capture their cumulative impact, assuming that quantization noise is additive.
\end{enumerate}

\paragraph{Bit-width allocation and budget selection.}
The aggregated sensitivities $D_k(b)$ and token counts $N_k$ are passed to the optimization algorithm (\Cref{sec:optim}) to produce the per-tag allocation $\textbf{b}^\star$ for a target average bit-width $B$.
We choose $B$ by sweeping $B\in [2, 4]$ on the calibration set and selecting the smallest value at which accuracy remains comparable to the BF16 baseline.
Since both $D_k(b)$ and $N_k$ are pre-computed, re-running bit-width allocation is highly efficient.

\subsection{System Design and Implementation}

This section describes how TriAxialKV is implemented in a serving system: the quantization scheme and buffer layout, the mixed-precision INT2 and INT4 memory pools, and the request lifecycle, which applies the per-tag allocation from~\Cref{sec:allocation} and runs the prefill and fused decode kernels.

\begin{figure}
    \centering
    \includegraphics[width=1\linewidth]{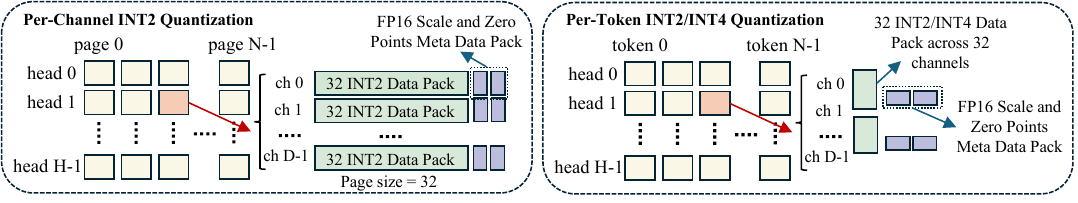}
    \caption{KV cache buffer layouts. Left: per-channel INT2 keys, indexed by (page, head). Right: per-token INT2 values and INT4 KV, indexed by (token, head). Page size matches the quantization group size G = 32. Quantized data are packed into unsigned INT8 buffers.}
    \label{fig:quantization}
\end{figure}

\subsubsection{Quantization Scheme and Buffer Layout}
\label{sec:quant_scheme}
\paragraph{Bit-widths and quantization axes.}
\Cref{fig:quantization} shows our quantization scheme and packed buffer layout.
We use an asymmetric group-wise quantization scheme with group size $G=32$ throughout.
For INT4, we quantize \emph{both keys and values per-token}.
Channels in a head are split into groups of $G$ elements, each quantized independently with its own (scale, zero-point) pair.
For INT2, we \emph{quantize keys per-channel and values per-token}.
Keys in a quantization group share one (scale, zero-point) pair \emph{for each channel}.
We quantize INT2 keys per-channel because, at low bit-width, high-magnitude outliers in some channels inflate the scale and destroy resolution across other channels under per-token quantization~\citep{liu2024kivi}.
Values are well-behaved under per-token quantization at both bit-widths.

\paragraph{Buffer layout.}
The INT2 key buffer, which uses per-channel quantization, is indexed by \texttt{(page, head)}.
As shown in \Cref{fig:quantization}, each entry holds the quantized keys for all $G$ tokens of the page under that head, grouped by channel, along with scales and zero-points.
This ensures that each channel's keys for the $G$ tokens are contiguous in memory, enabling efficient coalesced loads.
For per-token INT4 KV and INT2 values, each buffer is instead indexed by \texttt{(token, head)}.
Each entry stores the quantized data for all channels under one \texttt{(token, head)}, followed by their scales and zero-points.
Quantized data are stored as unsigned INT8 buffers with four INT2 or two INT4 values packed per byte, whereas scales and zero-points are stored in FP16.

\paragraph{Residual policy for INT2 keys.}
Per-channel quantization of INT2 keys requires the scale and zero-point of a channel to be shared across a group of $G$ tokens.
For the residual tokens that do not form a group, we \emph{route them through the INT4 per-token path} instead.
In KIVI~\citep{liu2024kivi}, the residual tokens are stored in full precision in a dedicated buffer, but we do not take this approach for two reasons.
First, adding a residual buffer complicates memory allocation and forces the attention kernel to handle three precisions.
Second, KIVI's residual buffer holds up to 128 tokens' KV in full precision per request in every layer.
This adds considerable overhead and erodes the memory savings that our INT2 path is meant to deliver, especially when there are many short requests.
Keeping the residual tokens on the INT4 path makes memory management simple and reduces overhead.

\subsubsection{Mixed-Precision Memory Pool}
\paragraph{Shared address space.}
Both the INT2 and INT4 memory pools \emph{share a single virtual address space}, split by an offset.
The offset is set at startup from the calibrated average bit-width $B$, which sets the ratio of INT2 and INT4 tokens.
Addresses below the offset index the INT2 pool, and addresses at or above the offset index the INT4 pool.
Downstream components, such as the attention backend, can infer a token's precision by comparing its memory address against the offset.

\paragraph{Memory allocation.}
Each pool has its own page allocator for memory allocation and freeing.
When a request enters the scheduler, the bit-width tagger produces a per-token bit-width array.
INT2 and INT4 tokens are routed to their respective allocators for memory allocation, and each allocator returns free page indices from its own list.
No cross-pool coordination is required.

\subsubsection{Request Lifecycle}
\label{sec:request_lifecycle}

\paragraph{Tagging.}
At serving time, each incoming request is tagged on the CPU during request scheduling and \emph{assigned an array of per-token bit-widths} that propagates to the downstream memory allocator and attention backend.
Specifically, we run the triaxial tagger from \Cref{sec:allocation} on the tokenized input to produce a per-token tag code.
Each tag code is looked up in the bit-width allocation map $\textbf{b}^\star:\mathcal{S} \to \{2, 4\}$ from the calibration, yielding a per-token bit-width array attached to the request.
To keep the tagger general across models and workloads, we drive it entirely by chat-template structure.
We use special-token IDs from the tokenizer, such as role markers, message delimiters, and structural brackets for thinking and tool calls, to segment the sequence with a single scan.

\paragraph{Prefill.}
For prefill, we implement a Triton~\citep{tillet2019triton} gather-dequant kernel that reads the matched-prefix slot indices from the INT2 and INT4 pools, unpacks each slot's payload, dequantizes it into the activation precision (BF16), and writes the results into a buffer.
The buffer is then passed to the FlashInfer~\citep{ye2025flashinfer} prefill attention kernel without further modification.
Queries, keys, and values for the newly computed tokens are passed directly to the prefill kernel alongside the dequantized prefix.

\paragraph{Decode.}
\begin{figure}
    \centering
    \includegraphics[width=1\linewidth]{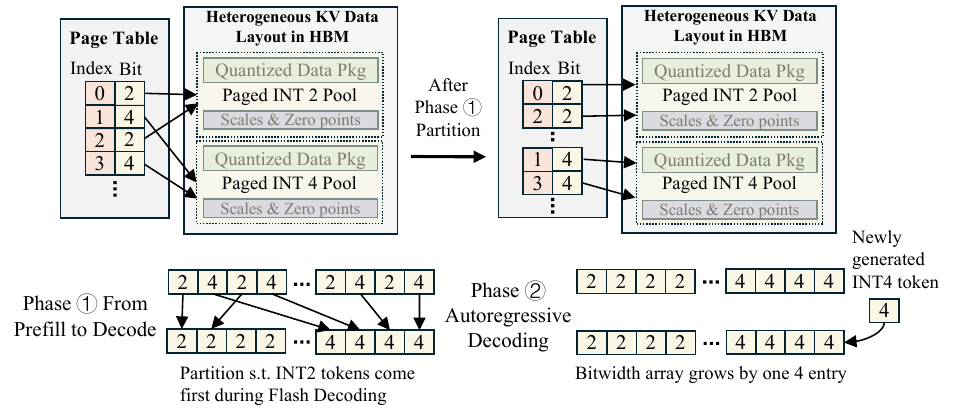}
    \caption{Per-request page table management. Phase 1 partitions page table entries so INT2 pointers precede INT4, enabling bit-width-homogeneous flash-decoding splits; Phase 2 generates INT4 tokens autoregressively during decode and grows the page table by one entry per step.}
    \label{fig:memory_management}
\end{figure}

After prefill, the decode stage generates new tokens.
As shown in \Cref{fig:memory_management}, each request's KV state is tracked in a per-request page table that records per-token pointers to the memory pool.
In phase 1, we \emph{partition} the pointers so that all INT2 pointers precede the INT4 pointers.
This suits our custom fused decode kernel, which follows Flash-Decoding~\citep{dao2023flashdecoding}: the KV is processed in parallel contiguous splits whose partial outputs are merged with a log-sum-exp reduction.
Our partition ensures that the leading splits handle INT2 KV and the trailing ones handle INT4.
Each split is bit-width-homogeneous and runs a single dequantization path.
In phase 2, each autoregressive decode step generates a new token, whose KV is stored in INT4.
Our fused dequantization-attention kernel has two paths: INT2 and INT4.
Each path loads only the required KV data, unpacks it via bit shifts, and applies dequantization on the fly during attention computation.
Keys and values of the generated token are written to the INT4 pool, and its address is appended to the page table.

\section{Experimental Results}
\begin{table}[t]
\centering
\small
\caption{Task accuracy (\%) on BFCL Memory, reported as mean and standard deviation over three runs. TriAxialKV Mixed stores 75\% of prefill tokens at INT2 (70\% for Qwen3-14B).}
\begin{tabular*}{\linewidth}{@{\extracolsep{\fill}}llllll}\toprule
\textbf{Method} & \textbf{Falcon3-10B-} & \textbf{Qwen3-14B} & \textbf{Qwen3-32B} & \textbf{Qwen3-235B-} \\
& \textbf{Instruct} &  &  & \textbf{A22B-FP8} \\
\midrule
SGLang BF16~\citep{zheng2024sglang} & 25.04\std{1.26} & 25.41\std{1.36} & 25.19\std{1.22} & 23.78\std{0.89} \\
SGLang FP4~\citep{zheng2024sglang} & 26.37\std{2.65} & 21.19\std{2.76} & 20.52\std{0.34} & 20.96\std{0.46}  \\
KIVI~\citep{liu2024kivi} & 22.52\std{0.46} & 19.41\std{1.99} & 20.74\std{1.23} & 20.37\std{2.06} \\
OSCAR~\citep{zhou2026oscarofflinespectralcovarianceaware} & 21.33\std{1.68} & 9.34\std{1.60}$^\dagger$ & 24.30\std{0.71} & 18.52\std{3.04} \\
\midrule
TriAxialKV INT4 & 24.30\std{2.34} & 22.15\std{2.38} & 23.93\std{0.93} & 24.44\std{1.18} \\
TriAxialKV Mixed & 23.04\std{2.27} & 23.48\std{2.45} & 24.00\std{1.11} & 22.59\std{1.28} \\
\bottomrule
\end{tabular*}
\par\vspace{2pt}
{\footnotesize\raggedright $^\dagger$Affected by a head-indexing bug in OSCAR's decode kernel that is triggered by Qwen3-14B's attention-head configuration.\par}
\vspace{-3mm}
\label{tab:bfcl_accuracy}
\end{table}

\subsection{Experimental Setup}
\label{sec:experimental_setup}
\paragraph{Benchmarks and models.}
We evaluate TriAxialKV on the memory category of the Berkeley Function Calling Leaderboard (BFCL)~\citep{patil2025bfcl} and OSWorld~\citep{OSWorld}, which cover two representative agentic regimes: text-only function calling with structured tool schemas, and multimodal computer-use trajectories with screenshot observations.
On BFCL, we evaluate Qwen3~\citep{yang2025qwen3} (14B, 32B, and 235B-A22B-Instruct-2507-FP8) and Falcon3-10B-Instruct~\citep{Falcon3}.
On OSWorld, we evaluate Qwen3-VL~\citep{Qwen3-VL} (8B-Thinking, 32B-Thinking) and InternVL3.5-38B~\citep{wang2025internvl3_5}.%
\footnote{The licenses for all models, benchmarks, and libraries used in this work are listed in Appendix~\ref{sec:license}.}
For the axis ablation, we additionally evaluate Qwen3.5-4B~\citep{qwen3.5} on Claw-Eval~\citep{ye2026claw}, a multimodal agentic benchmark.

\paragraph{Baselines.} We compare against four baselines covering the relevant prior art.
SGLang~\citep{zheng2024sglang} BF16 serves the model with full-precision KV cache and is the lossless reference.
SGLang FP4 uses SGLang's built-in FP4 KV cache backend, representing uniform low-bit floating-point quantization.
We further compare with two 2-bit KV cache quantization methods: KIVI~\citep{liu2024kivi}, a seminal asymmetric scheme, and OSCAR~\citep{zhou2026oscarofflinespectralcovarianceaware}, a recent rotation-based scheme.
We also report TriAxialKV INT4, which stores all tokens at INT4, to isolate the effect of the per-tag INT2/INT4 allocation.

\paragraph{End-to-end serving performance.}
We measure end-to-end serving throughput in tokens per second (TPS) on Qwen3-VL-8B and Qwen3-VL-32B, running OSWorld trajectories.
The throughput is measured on both a single NVIDIA B200 GPU (180 GB HBM3e) and a single NVIDIA H100 GPU (80 GB HBM3), both hosted in our internal cluster.
For each (model, GPU) configuration, we measure the throughput at the largest possible batch size without running out of memory.

\subsection{Task Accuracy}

\begin{figure}[t]
\centering
\begin{minipage}[t]{0.62\textwidth}
  \vspace{0pt}
  \centering
  \captionof{table}{Task accuracy (\%) on OSWorld across multimodal models. Gray numbers in brackets indicate deltas relative to the BF16 baseline. TriAxialKV Mixed maintains the accuracy of the BF16 reference, with a substantially lower memory budget.}
  \resizebox{\linewidth}{!}{%
\begin{tabular}{llll}
\toprule
\textbf{Method} & \textbf{Qwen3-VL-8B-} & \textbf{Qwen3-VL-32B-} & \textbf{InternVL3.5-} \\
& \textbf{Thinking} & \textbf{Thinking} & \textbf{38B}\\
\midrule
SGLang BF16~\citep{zheng2024sglang} & 35.70 \delt{+0.00} & 39.69 \delt{+0.00} & 12.22 \delt{+0.00} \\
SGLang FP4~\citep{zheng2024sglang} & 29.52 \delt{-6.18} & 40.54 \delt{+0.85} & 12.13 \delt{-0.09} \\
KIVI~\citep{liu2024kivi} & 34.61 \delt{-1.09} & 41.49 \delt{+1.80} & 12.27 \delt{+0.05} \\
OSCAR~\citep{zhou2026oscarofflinespectralcovarianceaware} & 30.52 \delt{-5.18} & 40.20 \delt{+0.51} & 11.95 \delt{-0.27} \\
\midrule
TriAxialKV INT4 & 35.38 \delt{-0.32} & 41.43 \delt{+1.74} & 13.11 \delt{+0.89} \\
TriAxialKV Mixed & 35.78 \delt{+0.08} & 40.59 \delt{+0.90} & 12.77 \delt{+0.55} \\
\bottomrule
\end{tabular}
}

  \label{tab:osworld_accuracy}
\end{minipage}
\hfill
\begin{minipage}[t]{0.37\textwidth}
  \vspace{-6pt}
  \centering
  \includegraphics[width=\linewidth]{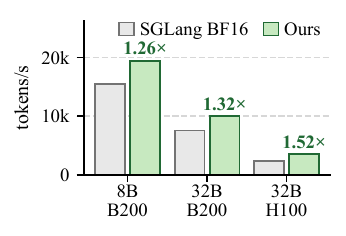}
  \vspace{-8mm}
  \captionof{figure}{End-to-end throughput on OSWorld trajectories. TriAxialKV delivers 1.26--1.52$\times$ the throughput of SGLang BF16.}
  \label{fig:throughput_comparison}
\end{minipage}
\end{figure}

Tables~\ref{tab:bfcl_accuracy} and~\ref{tab:osworld_accuracy} report task accuracy on BFCL Memory and OSWorld.
Across all experiments, our mixed-precision allocation \emph{tracks the BF16 reference closely}.
On BFCL Memory, with 75\% of prefill tokens at INT2 (70\% on Qwen3-14B), TriAxialKV stays within 2.0\% of the BF16 baseline on every model.
On OSWorld, with 85\% of prefill tokens at INT2, TriAxialKV matches the BF16 accuracy on every model.
TriAxialKV Mixed also \emph{performs on par with INT4}, showing that moving most tokens to INT2 under the calibrated allocation costs little accuracy.
Appendix~\ref{sec:accuracy_breakdown} gives per-category accuracy and OSWorld confidence intervals, and Appendix~\ref{sec:output_length} gives output lengths.

The quantization baselines are less consistent across models.
On BFCL Memory, SGLang FP4 matches BF16 on Falcon3-10B but drops 4--5 points on Qwen3-14B and Qwen3-32B.
KIVI falls below BF16 on every model by 2.5\%--6.0\%, and OSCAR has a competitive accuracy only on Qwen3-32B.
On OSWorld, Qwen3-VL-8B is notably more sensitive to KV quantization: SGLang FP4 and OSCAR degrade accuracy by 6.2\% and 5.2\%, and KIVI by 1.1\%.
Meanwhile, TriAxialKV stays close to BF16 on every model across both benchmarks.

Comparing TriAxialKV Mixed against the two 2-bit baselines, KIVI and OSCAR, \emph{demonstrates the value of per-tag allocation}.
Both baselines apply 2-bit quantization to all non-residual tokens, including the system prompt and tool schemas, which our calibrated allocation keeps at INT4.
These tokens carry the function signatures and argument specifications that the model \emph{must reproduce verbatim in its tool calls}, so quantization errors in their KV can plausibly propagate into incorrect argument names or types, which BFCL Memory scores as task failures.
TriAxialKV Mixed spends the 4-bit budget on exactly this $\mathtt{inst}$-tagged segment while pushing less sensitive segments to 2-bit, recovering most of the accuracy that uniform 2-bit quantization gives up.

\subsection{Serving Throughput}
\label{sec:throughput}
\begin{table}[t]
\centering
\small
\caption{End-to-end serving comparison on Qwen3-VL-32B-Thinking, measured on a single NVIDIA B200 GPU by replaying OSWorld trajectories. A more comprehensive comparison, including bytes/token, residual windows, and absolute capacity, is provided in Appendix~\ref{sec:throughput_full}.}
\begin{tabular*}{\linewidth}{@{\extracolsep{\fill}}lllrrrr}
\toprule
\textbf{Method} & \textbf{Prefill} & \textbf{Decode} & \textbf{Bits/} & \textbf{Capacity} & \textbf{Throughput} & \textbf{Speedup} \\
 & & & \textbf{elem.} & \textbf{vs.\ BF16} & \textbf{(tokens/s)} & \\
\midrule
SGLang BF16 & FlashInfer & SGLang Triton & 16.0 & 1.00$\times$ & 7,787 & 1.00$\times$ \\
SGLang BF16 & SGLang Triton & SGLang Triton & 16.0 & 1.00$\times$ & 5,983 & 0.77$\times$ \\
SGLang FP4 & TRT-LLM & TRT-LLM & 4.5 & 3.16$\times$ & 4,287 & 0.55$\times$ \\
KIVI$^\dagger$ & SGLang Triton & SGLang Triton & 3.0 & 1.00$\times$ & 5,772 & 0.74$\times$ \\
OSCAR & FA-4 & OSCAR Triton & 2.5 & 5.70$\times$ & 7,630 & 0.98$\times$ \\
\midrule
TriAxialKV INT4 & FlashInfer & Ours Triton & 5.0 & 3.18$\times$ & 9,407 & 1.21$\times$ \\
TriAxialKV Mixed & FlashInfer & Ours Triton & 3.3 & 4.51$\times$ & \textbf{10,274} & \textbf{1.32$\times$} \\
\bottomrule
\end{tabular*}
\par\vspace{2pt}
{\footnotesize\raggedright $^\dagger$KIVI has no official serving-system integration. We reimplemented it on SGLang with fake quantization to measure its task accuracy, so its compressed footprint yields neither capacity nor bandwidth gains.\par}
\label{tab:throughput_baselines}
\end{table}

To measure the serving throughput, we replay OSWorld trajectories, which have an average prefill length of 11,000 tokens and an average decode length of 300 tokens.
We compare TriAxialKV with the SGLang BF16 baseline.
Both methods use FlashInfer~\citep{ye2025flashinfer} for prefill; the SGLang BF16 baseline uses SGLang's original Triton decode kernel, while our method uses our custom fused Triton kernel for mixed-precision dequantization and attention.
Figure~\ref{fig:throughput_comparison} reports end-to-end throughput at the maximum concurrency each method can sustain.
On a single B200 GPU, TriAxialKV delivers \emph{1.32$\times$ the BF16 throughput} on Qwen3-VL-32B, and \emph{1.26$\times$} on Qwen3-VL-8B.
On a single H100 GPU, the gain on Qwen3-VL-32B rises to \emph{1.52$\times$}.
It also supports 3.4--4.0$\times$ as many concurrent requests.

We also compared TriAxialKV with other baselines on a single B200 GPU.
As shown in Table~\ref{tab:throughput_baselines}, TriAxialKV INT4 and SGLang FP4 have a similar capacity, but TriAxialKV INT4 delivers 2.2$\times$ the throughput of the TRT-LLM FP4 backend in SGLang.
Moving from INT4 to mixed allocation, TriAxialKV has a further 1.4$\times$ increase in capacity and increases the speedup to 1.32$\times$.
OSCAR holds the largest KV cache capacity but shows no end-to-end speedup.
In OSCAR's implementation, each prefill rotates the QKV, dequantizes the cached prefix to BF16, and assembles each request's KV in a Python loop that forces GPU--CPU synchronization in every layer.
We attribute the missing speedup to this overhead, which offsets the capacity gain on this prefill-heavy workload.

\subsection{Ablation Study}
\begin{figure}[t]
\centering
\begin{minipage}[t]{0.59\textwidth}
  \centering
  \captionof{table}{Drop-one-axis ablation. Removing any applicable axis degrades task performance.}
  \resizebox{\linewidth}{!}{%
\begin{tabular}{lccc}
\toprule
 & \multicolumn{2}{c}{\textbf{BFCL Memory}} & \multicolumn{1}{c}{\textbf{Claw-Eval}} \\
\cmidrule(lr){2-3} \cmidrule(lr){4-4}
\textbf{Variant} & Qwen3-14B & Qwen3-32B & Qwen3.5-4B \\
\midrule
No Temporal   & 22.00 & 24.00 & 31.60 \\
No Modal      & --    & --    & 32.18 \\
No Semantic   & 18.00 & 20.89 & 33.10 \\
Full          & \textbf{24.22} & \textbf{25.11} & \textbf{34.17} \\
\bottomrule
\end{tabular}
}

  \label{tab:ablation_axes}
\end{minipage}
\hfill
\begin{minipage}[t]{0.40\textwidth}
  \centering
  \captionof{table}{Memory budget sweep for Qwen3-14B on BFCL Memory.}

\resizebox{\linewidth}{!}{%
\begin{tabular}{lcc}
\toprule
INT2 Tokens & Accuracy (\%) & $\Delta$ vs.\ BF16 \\
\midrule
100\% & 19.56 & $-5.85$ \\
90\% & 20.37 & $-5.04$ \\
80\% & 20.08 & $-5.33$ \\
75\% & 21.63 & $-3.78$ \\
70\% & \textbf{23.48} & $-1.93$ \\
\midrule
BF16 & 25.41 & 0.00 \\
\bottomrule
\end{tabular}
}

  \label{tab:ablation_budget}
\end{minipage}
\end{figure}

\paragraph{Axis ablation.}
\Cref{tab:ablation_axes} removes one axis of the taxonomy at a time, on BFCL Memory and Claw-Eval.
Dropping an axis merges all tags that differ only along that axis into one, and we re-run calibration and allocation on the reduced tag space at the same average bit-width.
\emph{No Temporal} treats all tokens as belonging to a single turn, removing temporal recency from the tag space.
\emph{No Modal} treats image and text tokens as the same.
\emph{No Semantic} treats all tokens as a single semantic role, regardless of whether they are system prompts, user queries, or tool calls.

Removing any axis lowers accuracy, but which axis matters most depends on the workload.
On BFCL Memory, the semantic axis dominates: removing it leads to a degradation of 6.22\% on Qwen3-14B and 4.22\% on Qwen3-32B, three to four times the cost of removing the temporal axis.
The semantic axis gives the system prompt and tool schemas their own tags, so the allocator can keep them at INT4.
Errors in these segments propagate into incorrect function and argument names, which cause the task to fail.
Without the semantic axis, these tokens are merged with other roles that share their temporal and modality tags, and the merged tag is assigned INT2.
On Claw-Eval, the order reverses: removing temporal recency costs the most, followed by modality and semantic role.
The dominant axis shifts with the workload, which is why the taxonomy keeps all three rather than specializing to one.

\paragraph{Memory budget.}

Table~\ref{tab:ablation_budget} sweeps the share of prefill tokens stored at INT2 on Qwen3-14B (BFCL Memory), from uniform INT2 to 70\%.
Moving the first 20\% of tokens to INT4 leaves it essentially unchanged, while the next 10\% recovers most of the gap, reaching 23.48\%.
Because the budget at which accuracy recovers differs across models and workloads, a single fixed budget either wastes memory or falls short of it, which is why we calibrate it per model and workload.

\section{Conclusion}
We presented TriAxialKV, a mixed-precision KV-cache quantization framework that exploits the structure of agentic prefills along three complementary axes: temporal recency, modality, and semantic role.
TriAxialKV tags each prefill token using only the chat template, calibrates the attention-output distortion of each tag, and allocates INT2/INT4 bit-widths under a memory budget.
It runs in SGLang with custom paged mixed-precision memory pools and fused Triton decode kernels.
On BFCL Memory and OSWorld, TriAxialKV preserves the task accuracy relative to the BF16 reference.
On Qwen3-VL-32B (OSWorld), it provides 4.5$\times$ KV-cache capacity and delivers 1.32$\times$ the BF16 throughput on B200 and 1.52$\times$ on H100.
These results show that jointly reasoning about temporal, modal, and semantic structure is both necessary and practical for aggressive KV-cache compression.

\begin{ack}
We thank Taowen Liu and Can Xiao for their helpful discussions.
This work was supported by the Advanced Research and Invention Agency (ARIA) under the \href{https://aria.org.uk/opportunity-spaces/nature-computes-better/scaling-compute/}{Scaling Compute Program}.
\end{ack}

\bibliography{main}
\bibliographystyle{plain}

\newpage
\appendix
\section{Greedy Bit-Width Allocation}
\label{sec:greedy_allocation}
\begin{algorithm}[h]
\caption{Triaxial bit-width allocation}
\label{alg:alloc}
\begin{algorithmic}[1]
  \Require Tag set $\mathcal{S}$; counts $\{N_s\}$; distortions $\{D_s(2), D_s(4)\}$; budget $B$.
  \Ensure Allocation $\mathbf{b}^\star \in \{2,4\}^{|\mathcal{S}|}$.
  \State Compute $\rho_s \gets (D_s(2) - D_s(4)) \,/\, (2 N_s)$ for all $s \in \mathcal{S}$.
  \State Sort $\mathcal{S}$ by $\rho_s$ descending; set $b_s \gets 2$ and $R \gets (B - 2) \sum_{s} N_s$.
  \For{$s$ in sorted order}
    \If{$2 N_s \le R$}
      \State $b_s \gets 4$; \, $R \gets R - 2 N_s$.
    \EndIf
  \EndFor
  \State \Return $\mathbf{b}^\star = (b_s)_{s \in \mathcal{S}}$.
\end{algorithmic}
\end{algorithm}

\section{Baseline Selection}
\label{sec:baseline_selection}
Agentic benchmarks run multi-turn episodes against a live serving engine, so a KV-cache method needs a serving integration to be evaluated on them, and efficient kernels for a throughput comparison to be meaningful.
\Cref{tab:baseline_features} lists the candidates we considered.
We select SGLang FP4, the built-in low-bit KV cache of the framework we build on; OSCAR, the only 2-bit method in the table that is both open source and integrated with a serving system; and KIVI, a widely used asymmetric 2-bit scheme, which we reimplement in SGLang with fake quantization to measure its accuracy.
KVQuant, PM-KVQ, and LOOK-M release code without a serving integration, and ThinKV and VL-Cache have no public code.
\begin{table}[!htb]
\centering
\caption{KV-cache compression methods considered as baselines. \emph{Serving integration}: runs inside a serving engine, which agentic benchmarks require for end-to-end evaluation. \emph{End-to-end speedup}: the original work reports serving-throughput gains. \emph{Residual buffer}: keeps a full-precision window of tokens per request, whose memory does not shrink with the bit-width.}
\begin{tabular*}{\linewidth}{@{\extracolsep{\fill}}lcccc}
\toprule
\textbf{Method} & \textbf{Open} & \textbf{Serving} & \textbf{End-to-end} & \textbf{Residual} \\
 & \textbf{source} & \textbf{integration} & \textbf{speedup} & \textbf{buffer} \\
\midrule
SGLang FP4~\citep{zheng2024sglang} & \cmark & \cmark & \xmark & \xmark \\
KIVI~\citep{liu2024kivi} & \cmark & \xmark & \xmark & \cmark \\
KVQuant~\citep{hooper2024kvquant} & \cmark & \xmark & \xmark & \xmark \\
QServe~\citep{lin2025qserve} & \cmark & \cmark & \cmark & \xmark \\
PM-KVQ~\citep{liu2026pmkvq} & \cmark & \xmark & \xmark & \xmark \\
ThinKV~\citep{thinkv} & \xmark & \cmark & \cmark & \cmark \\
VL-Cache~\citep{tu2025vlcache} & \xmark & \cmark & \cmark & \xmark \\
LOOK-M~\citep{wan2024lookm} & \cmark & \xmark & \xmark & \xmark \\
OSCAR~\citep{zhou2026oscarofflinespectralcovarianceaware} & \cmark & \cmark & \cmark & \cmark \\
\midrule
TriAxialKV (Ours) & \cmark & \cmark & \cmark & \xmark \\
\bottomrule
\end{tabular*}
\label{tab:baseline_features}
\end{table}

\section{Per-Category Accuracy on BFCL and OSWorld}
BFCL Memory evaluates whether an agent can store and recall information across sessions with one of three memory backends: a key-value store (KV), a vector store (Vector), and recursive summarization (RecSum).
OSWorld groups its tasks by application domain: OS (operating-system and terminal tasks), Office (LibreOffice Calc, Writer, and Impress), Daily (Chrome, VLC, and Thunderbird), Professional (VS Code and GIMP), and Workflow (tasks that span several applications).
Several categories are small; OS, for example, has about two dozen tasks, so a single task moves its accuracy by more than 4 points.
\Cref{tab:osworld_ci} therefore also reports 95\% bootstrap confidence intervals for overall OSWorld accuracy.
\label{sec:accuracy_breakdown}

\begin{table}[H]                            
  \centering
  \small
  \caption{Per-category accuracy (\%) on BFCL Memory across the KV, Vector, and RecSum sub-tasks, averaged   
  over the same three runs as Table~\ref{tab:bfcl_accuracy}; Overall is the mean of the three sub-tasks.}    
  \begin{tabular*}{\linewidth}{@{\extracolsep{\fill}}lcccc}                                                  
  \toprule                                                                                                   
  \textbf{Method} & \textbf{Overall} & \textbf{KV} & \textbf{Vector} & \textbf{RecSum} \\                    
  \midrule                                                                                                   
  \multicolumn{5}{l}{\textit{Falcon3-10B-Instruct}} \\                                                       
  SGLang BF16~\citep{zheng2024sglang} & 25.04 & 10.45 & 16.22 & 48.44 \\                                     
  SGLang FP4~\citep{zheng2024sglang} & 26.37 & 12.89 & 16.45 & 49.78 \\                                      
  KIVI~\citep{liu2024kivi} & 22.52 & 11.78 & 18.00 & 37.78 \\                                                
  OSCAR~\citep{zhou2026oscarofflinespectralcovarianceaware} & 21.33 & 11.33 & 17.33 & 35.33 \\               
  TriAxialKV INT4 & 24.30 & 10.89 & 16.45 & 45.56 \\                                                         
  TriAxialKV Mixed & 23.04 & 12.22 & 15.11 & 41.78 \\                                                        
  \midrule                                                                                                   
  \multicolumn{5}{l}{\textit{Qwen3-14B}} \\                                                                  
  SGLang BF16~\citep{zheng2024sglang} & 25.41 & 12.00 & 18.89 & 45.33 \\                                     
  SGLang FP4~\citep{zheng2024sglang} & 21.19 & 9.33 & 11.78 & 42.44 \\                                       
  KIVI~\citep{liu2024kivi} & 19.41 & 5.33 & 14.00 & 38.89 \\                                                 
  OSCAR~\citep{zhou2026oscarofflinespectralcovarianceaware} & 9.34 & 8.44 & 8.22 & 11.33 \\                  
  TriAxialKV INT4 & 22.15 & 8.44 & 14.67 & 43.33 \\                                                          
  TriAxialKV Mixed & 23.48 & 11.78 & 13.78 & 44.89 \\
  \midrule
  \multicolumn{5}{l}{\textit{Qwen3-32B}} \\
  SGLang BF16~\citep{zheng2024sglang} & 25.19 & 16.22 & 15.11 & 44.22 \\
  SGLang FP4~\citep{zheng2024sglang} & 20.52 & 11.11 & 16.00 & 34.44 \\
  KIVI~\citep{liu2024kivi} & 20.74 & 12.67 & 15.78 & 33.78 \\
  OSCAR~\citep{zhou2026oscarofflinespectralcovarianceaware} & 24.30 & 13.78 & 17.56 & 41.56 \\
  TriAxialKV INT4 & 23.93 & 14.67 & 16.44 & 40.66 \\
  TriAxialKV Mixed & 24.00 & 11.11 & 20.22 & 40.67 \\
  \midrule
  \multicolumn{5}{l}{\textit{Qwen3-235B-A22B-Instruct-2507-FP8}} \\
  SGLang BF16~\citep{zheng2024sglang} & 23.78 & 9.56 & 16.67 & 45.11 \\
  SGLang FP4~\citep{zheng2024sglang} & 20.96 & 7.56 & 11.11 & 44.22 \\
  KIVI~\citep{liu2024kivi} & 20.37 & 11.78 & 12.89 & 36.44 \\
  OSCAR~\citep{zhou2026oscarofflinespectralcovarianceaware} & 18.52 & 10.22 & 14.89 & 30.44 \\
  TriAxialKV INT4 & 24.44 & 13.33 & 14.22 & 45.78 \\
  TriAxialKV Mixed & 22.59 & 10.00 & 14.45 & 43.33 \\
  \bottomrule
  \end{tabular*}
  \label{tab:bfcl_breakdown}
  \end{table}

\begin{table}[H]
\centering
\small
\caption{Per-category accuracy (\%) on OSWorld by application category.}
\begin{tabular*}{\linewidth}{@{\extracolsep{\fill}}lcccccc}
\toprule
\textbf{Method} & \textbf{Overall} & \textbf{OS} & \textbf{Office} & \textbf{Daily} & \textbf{Professional} & \textbf{Workflow} \\
\midrule
\multicolumn{7}{l}{\textit{Qwen3-VL-8B-Thinking}} \\
SGLang BF16~\citep{zheng2024sglang} & 35.70 & 47.83 & 29.67 & 51.21 & 48.94 & 19.65 \\
SGLang FP4~\citep{zheng2024sglang} & 29.52 & 50.00 & 25.77 & 43.08 & 31.82 & 13.06 \\
KIVI~\citep{liu2024kivi} & 34.61 & 43.48 & 30.30 & 44.27 & 51.06 & 20.86 \\
OSCAR~\citep{zhou2026oscarofflinespectralcovarianceaware} & 30.52 & 34.78 & 26.94 & 46.90 & 42.55 & 13.73 \\
TriAxialKV INT4 & 35.38 & 39.13 & 33.28 & 50.41 & 48.94 & 17.34 \\
TriAxialKV Mixed & 35.78 & 43.48 & 32.97 & 52.72 & 40.43 & 21.00 \\
\midrule
\multicolumn{7}{l}{\textit{Qwen3-VL-32B-Thinking}} \\
SGLang BF16~\citep{zheng2024sglang} & 39.69 & 52.17 & 33.58 & 53.73 & 48.94 & 27.30 \\
SGLang FP4~\citep{zheng2024sglang} & 40.54 & 43.48 & 39.22 & 56.43 & 48.94 & 23.60 \\
KIVI~\citep{liu2024kivi} & 41.49 & 56.52 & 39.21 & 47.10 & 57.45 & 27.21 \\
OSCAR~\citep{zhou2026oscarofflinespectralcovarianceaware} & 40.20 & 54.55 & 38.34 & 48.92 & 48.94 & 26.68 \\
TriAxialKV INT4 & 41.43 & 52.17 & 44.59 & 49.51 & 48.94 & 23.94 \\
TriAxialKV Mixed & 40.59 & 43.48 & 37.41 & 52.59 & 53.19 & 27.35 \\
\midrule
\multicolumn{7}{l}{\textit{InternVL3.5-38B}} \\
SGLang BF16~\citep{zheng2024sglang} & 12.22 & 21.74 & 6.30 & 21.51 & 21.28 & 4.55 \\
SGLang FP4~\citep{zheng2024sglang} & 12.13 & 0.00 & 7.97 & 23.59 & 17.02 & 2.04 \\
KIVI~\citep{liu2024kivi} & 12.27 & 21.74 & 5.48 & 28.11 & 10.87 & 5.68 \\
OSCAR~\citep{zhou2026oscarofflinespectralcovarianceaware} & 11.95 & 17.39 & 9.93 & 21.57 & 12.77 & 4.55 \\
TriAxialKV INT4 & 13.11 & 23.81 & 5.40 & 24.22 & 21.28 & 6.10 \\
TriAxialKV Mixed & 12.77 & 26.09 & 6.27 & 24.20 & 19.15 & 4.55 \\
\bottomrule
\end{tabular*}
\label{tab:osworld_breakdown}
\end{table}

\begin{table}[H]
\centering
\small
\caption{OSWorld task accuracy (\%) with 95\% bootstrap confidence intervals in gray, obtained by resampling task-level results (10{,}000 resamples, percentile intervals).}
\begin{tabular*}{\linewidth}{@{\extracolsep{\fill}}llll}
\toprule
\textbf{Method} & \textbf{Qwen3-VL-8B-} & \textbf{Qwen3-VL-32B-} & \textbf{InternVL3.5-} \\
 & \textbf{Thinking} & \textbf{Thinking} & \textbf{38B} \\
\midrule
SGLang BF16      & 35.70 \delt{30.67--40.85} & 39.69 \delt{34.44--44.90} & 12.22 \delt{8.75--15.74} \\
SGLang FP4       & 29.52 \delt{24.58--34.74} & 40.54 \delt{35.37--45.69} & 12.13 \delt{8.49--16.03} \\
KIVI             & 34.61 \delt{29.72--39.68} & 41.49 \delt{36.32--46.74} & 12.27 \delt{8.81--15.84} \\
OSCAR            & 30.52 \delt{25.78--35.30} & 40.20 \delt{35.20--45.31} & 11.95 \delt{8.73--15.44} \\
\midrule
TriAxialKV INT4  & 35.38 \delt{30.30--40.55} & 41.43 \delt{36.30--46.55} & 13.11 \delt{9.55--16.69} \\
TriAxialKV Mixed & 35.78 \delt{30.70--40.99} & 40.59 \delt{35.39--45.72} & 12.77 \delt{9.30--16.50} \\
\bottomrule
\end{tabular*}
\label{tab:osworld_ci}
\end{table}

\section{Output Length}
\label{sec:output_length}
KV-cache quantization can change generation behavior even when accuracy is preserved, for example by causing repetition or cutting reasoning short.
\Cref{tab:output_length} reports the mean number of generated tokens per task on both benchmarks.
TriAxialKV Mixed stays within 3\% of the BF16 output length on the three Qwen3 models of BFCL Memory, and within 7\% on all OSWorld models.

\begin{table}[H]
\centering
\caption{Mean generated tokens per task. (a) BFCL Memory, mean $\pm$ standard deviation over three runs. (b) OSWorld.}
\begin{subtable}{\linewidth}
\centering
\caption{BFCL Memory.}
\begin{tabular*}{\linewidth}{@{\extracolsep{\fill}}lcccc}
\toprule
\textbf{Method} & \textbf{Falcon3-10B} & \textbf{Qwen3-14B} & \textbf{Qwen3-32B} & \textbf{Qwen3-235B} \\
\midrule
SGLang BF16 & 546$\pm$35 & 1601$\pm$70 & 2317$\pm$52 & 715$\pm$26 \\
SGLang FP4 & 657$\pm$40 & 1538$\pm$4 & 2353$\pm$20 & 822$\pm$31 \\
KIVI & 508$\pm$148 & 1690$\pm$29 & 2228$\pm$65 & 746$\pm$89 \\
OSCAR & 522$\pm$47 & 2189$\pm$131 & 2406$\pm$488 & 496$\pm$28 \\
TriAxialKV INT4 & 463$\pm$81 & 1612$\pm$39 & 2197$\pm$5 & 697$\pm$11 \\
TriAxialKV Mixed & 455$\pm$25 & 1636$\pm$41 & 2312$\pm$106 & 695$\pm$38 \\
\bottomrule
\end{tabular*}
\end{subtable}

\vspace{2mm}
\begin{subtable}{\linewidth}
\centering
\caption{OSWorld.}
\begin{tabular*}{\linewidth}{@{\extracolsep{\fill}}lccc}
\toprule
\textbf{Method} & \textbf{Qwen3-VL-8B} & \textbf{Qwen3-VL-32B} & \textbf{InternVL3.5-38B} \\
\midrule
SGLang BF16 & 10967 & 9351 & 3890 \\
SGLang FP4 & 11671 & 10357 & 3460 \\
KIVI & 11063 & 10206 & 4017 \\
OSCAR & 10533 & 9279 & 4851 \\
TriAxialKV INT4 & 9892 & 9542 & 3601 \\
TriAxialKV Mixed & 10196 & 9525 & 3874 \\
\bottomrule
\end{tabular*}
\end{subtable}
\label{tab:output_length}
\end{table}

\section{Full Serving Comparison}
\label{sec:throughput_full}
\Cref{tab:throughput_full} extends \Cref{tab:throughput_baselines} with per-token footprint, residual windows, and absolute capacity.
Bits per element include quantization metadata: with group size $G=32$ and an FP16 scale and zero-point per group, INT2 and INT4 cost 3 and 5 bits per element, so the 3.3 bits of TriAxialKV Mixed correspond to 85\% of tokens at INT2.
Bytes per token count keys and values across all layers and KV heads of Qwen3-VL-32B.
The residual column gives the full-precision tokens each method keeps per request: KIVI keeps a 128-token recent window and OSCAR keeps 256 sink and recent tokens, whereas TriAxialKV routes residual tokens through the INT4 path (\Cref{sec:quant_scheme}) and keeps none.

\begin{table}[H]
\centering
\caption{Serving comparison of all KV cache methods on Qwen3-VL-32B-Thinking (single B200, replayed OSWorld traces). Bits/element and bytes/token include quantization metadata and exclude the full-precision residual window, listed separately in tokens per request. Capacity is the number of tokens the KV cache pool holds.}
\resizebox{\linewidth}{!}{%
\begin{tabular}{llllllllll}
\toprule
\textbf{Method} & \textbf{Prefill} & \textbf{Decode} & \textbf{Bits/} & \textbf{Bytes/} & \textbf{Residual} & \textbf{Capacity} & \textbf{Capacity} & \textbf{Throughput} & \textbf{Speedup} \\
 & \textbf{backend} & \textbf{backend} & \textbf{element} & \textbf{token} & \textbf{(tokens/req)} & \textbf{(tokens)} & \textbf{vs.\ BF16} & \textbf{(tokens/s)} & \\
\midrule
SGLang BF16 & FlashInfer & SGLang Triton & 16.0 & 262,144 & -- & 325,952 & 1.00$\times$ & 7,786.74 & 1.00$\times$ \\
SGLang BF16 & SGLang Triton & SGLang Triton & 16.0 & 262,144 & -- & 326,003 & 1.00$\times$ & 5,982.67 & 0.77$\times$ \\
SGLang FP4 & TRT-LLM & TRT-LLM & 4.5 & 73,728 & 0 & 1,030,720 & 3.16$\times$ & 4,287.16 & 0.55$\times$ \\
KIVI$^\dagger$ & SGLang Triton & SGLang Triton & 3.0 & 49,152 & 128 & 326,003 & 1.00$\times$ & 5,771.76 & 0.74$\times$ \\
OSCAR & FlashAttention-4 & OSCAR Triton & 2.5 & 40,960 & 256 & 1,856,808 & 5.70$\times$ & 7,630.07 & 0.98$\times$ \\
\midrule
TriAxialKV INT4 & FlashInfer & Ours Triton & 5.0 & 81,920 & 0 & 1,037,063 & 3.18$\times$ & 9,407.18 & 1.21$\times$ \\
TriAxialKV Mixed & FlashInfer & Ours Triton & 3.3 & 54,067 & 0 & 1,469,312 & 4.51$\times$ & \textbf{10,274.45} & \textbf{1.32$\times$} \\
\bottomrule
\end{tabular}
}
\\[2pt]
{\footnotesize\raggedright $^\dagger$KIVI has no official serving-system integration. We reimplemented it on SGLang with fake quantization to measure its task accuracy, so its compressed footprint yields neither capacity nor bandwidth gains.\par}

\label{tab:throughput_full}
\end{table}

\section{Limitations}
\label{sec:limitations}
Our method has a few limitations:
\vspace{-1mm}
\begin{itemize}[leftmargin=1em]
    \item The per-tag sensitivity table is calibrated per workload and per model, requiring a one-time offline pass over a small calibration set. While inexpensive, this means the method is not zero-shot transferable to entirely new workloads.
    \vspace{-1mm}
    \item Our tagger relies on chat-template structure such as role markers, message delimiters, and brackets, and requires adaptation for models or prompting styles that lack standardized markers.
    \vspace{-1mm}
    \item Our allocator currently selects between INT2 and INT4 only. Supporting finer-grained bit-widths could yield further Pareto improvements, but would complicate kernel and memory-pool design.
\end{itemize}

\section{Broader Impacts}
\label{sec:broader_impacts}
TriAxialKV is a KV-cache compression technique that reduces the memory and energy cost of serving large language models.
We see two primary \textbf{positive} societal impacts.
First, lower memory consumption per request enables larger batch sizes and higher throughput on the same hardware, reducing the per-token energy cost of inference and the corresponding environmental footprint of large-scale LLM deployment.
Second, by allowing capable models to fit within the memory budget of less-expensive GPUs, our method lowers the hardware barrier to deploying agentic systems, broadening access for researchers and practitioners with limited compute resources.
As far as we are aware, our work does not introduce new negative societal impacts beyond those already inherent to the underlying models being served.
TriAxialKV does not modify model weights, training data, or output behavior; it is a serving-system optimization that preserves task accuracy relative to the BF16 reference. Any risks associated with the deployment of LLMs or agentic systems are inherited from the base models, and existing safeguards for those models continue to apply.

\section{Licenses for Existing Assets}
\label{sec:license}
We list below the existing assets used in this work, along with their licenses. All assets are used in accordance with their respective license terms.

\textbf{Models}:
\vspace{-1mm}
\begin{denseitemize}
    \item Qwen3 (14B, 32B, 235B-A22B-Instruct-2507-FP8)~\citep{yang2025qwen3}: Apache License 2.0.
    \item Qwen3-VL (8B-Thinking, 32B-Thinking)~\citep{Qwen3-VL}: Apache License 2.0.
    \item Qwen3.5-4B~\citep{qwen3.5}: Apache License 2.0.
    \item Falcon3-10B-Instruct~\citep{Falcon3}: TII Falcon-LLM License 2.0.
    \item InternVL3.5-38B~\citep{wang2025internvl3_5}: Apache License 2.0.
\end{denseitemize}

\textbf{Benchmarks}:
\vspace{-1mm}
\begin{denseitemize}
    \item Berkeley Function Calling Leaderboard (BFCL)~\citep{patil2025bfcl}: Apache License 2.0.
    \item OSWorld~\citep{OSWorld}: Apache License 2.0.
    \item Claw-Eval~\citep{ye2026claw}: MIT License.
\end{denseitemize}

\textbf{Serving frameworks and libraries}:
\vspace{-1mm}
\begin{denseitemize}
    \item SGLang~\citep{zheng2024sglang} (v0.5.15.post1): Apache License 2.0. Our implementation is built as a fork.
    \item FlashInfer~\citep{ye2025flashinfer}: Apache License 2.0. Used as the BF16 prefill attention backend.
    \item Triton~\citep{tillet2019triton}: MIT License. Used to implement our fused mixed-precision decode kernels.
    \item TensorRT-LLM~\citep{tensorrtllm}: Apache License 2.0. Provides the FP4 backend of the SGLang FP4 baseline.
    \item FlashAttention-4~\citep{zadouri2026flashattention4algorithmkernelpipelining}: BSD 3-Clause License. Prefill backend of the OSCAR baseline.
    \item KIVI~\citep{liu2024kivi}: MIT License. Reimplemented in SGLang as a baseline for comparison.
    \item OSCAR~\citep{zhou2026oscarofflinespectralcovarianceaware}: MIT License. Used as a baseline for comparison.
\end{denseitemize}

\end{document}